\documentclass{article}

\usepackage{amsmath}
\usepackage{amssymb}
\usepackage{authblk}
\usepackage[skip=0pt]{caption}
\usepackage{dblfloatfix}
\usepackage{graphicx}
\usepackage{hyperref}
\usepackage{natbib}
\usepackage{siunitx}
\usepackage{textcomp}
\usepackage{times}

\usepackage[accepted]{icml2017}

\icmltitlerunning{Deciding How to Decide: Dynamic Routing in Artificial Neural Networks}
\icmlkeywords{neural networks, reinforcement learning, vision, conditional computation, cascaded evaluation}

\begin{document} 
  \twocolumn[
    \icmltitle{Deciding How to Decide: \\ Dynamic Routing in Artificial Neural Networks}
    \begin{icmlauthorlist}
      \icmlauthor{Mason McGill}{caltech}
      \icmlauthor{Pietro Perona}{caltech}
    \end{icmlauthorlist}
    \icmlcorrespondingauthor{Mason McGill}{mmcgill@caltech.edu}
    \icmlaffiliation{caltech}{California Institute of Technology, Pasadena, California, USA}
    \vskip 0.3in
  ]
  \printAffiliationsAndNotice{}
  
  \begin{abstract}
    We propose and systematically evaluate three strategies for training dynamically-routed artificial neural networks: graphs of learned transformations through which different input signals may take different paths. Though some approaches have advantages over others, the resulting networks are often qualitatively similar. We find that, in dynamically-routed networks trained to classify images, layers and branches become specialized to process distinct categories of images. Additionally, given a fixed computational budget, dynamically-routed networks tend to perform better than comparable statically-routed networks.
  \end{abstract}

  \section{Introduction}
  
  Some decisions are easier to make than others---for example, large, unoccluded objects are easier to recognize. Additionally, different difficult decisions may require different expertise---an avid birder may know very little about identifying cars. We hypothesize that complex decision-making tasks like visual classification can be meaningfully divided into specialized subtasks, and that a system designed to perform a complex task should first attempt to identify the subtask being presented to it, then use that information to select the most suitable algorithm for its solution.
   
  This approach---dynamically routing signals through an inference system, based on their content---has already been incorporated into machine vision pipelines via methods such as boosting \cite{viola2005detecting}, coarse-to-fine cascades \cite{zhou2013extensive}, and random decision forests \cite{ho1995random}. Dynamic routing is also performed in the primate visual system: spatial information is processed somewhat separately from object identity information \cite{goodale1992separate}, and faces and other behaviorally-relevant stimuli ellicit responses in anatomically distinct, specialized regions \cite{moeller2008patches,kornblith2013network}. However, state-of-the-art artificial neural networks (ANNs) for visual inference are routed statically \cite{simonyan2014very,he2016deep,dosovitskiy2015flownet,newell2016stacked}; every input triggers an identical sequence of operations.
  
  \begin{figure}[htb]
    \centering
    \includegraphics[width=3in]{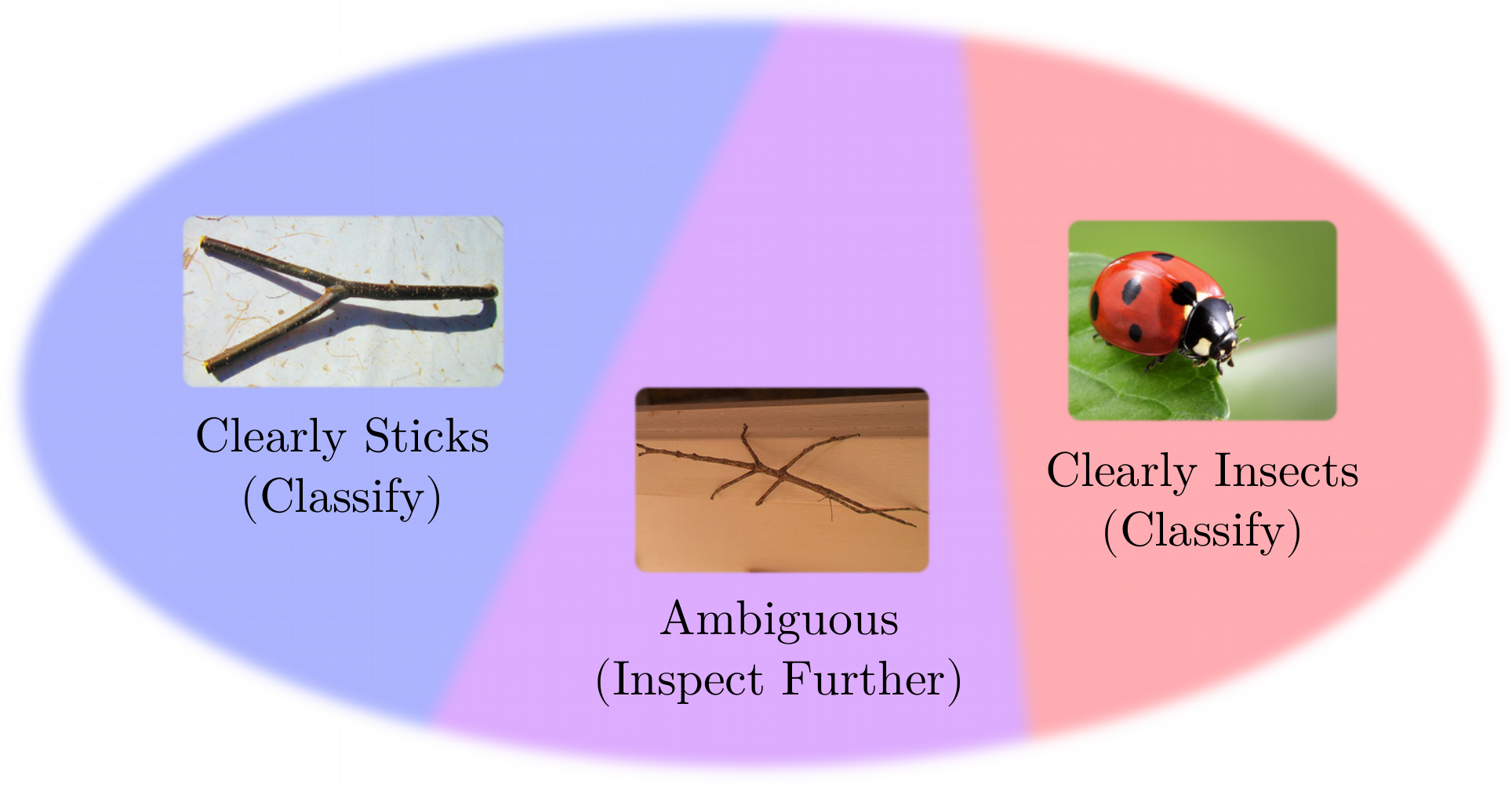}
    \caption{\textbf{Motivation for dynamic routing.} For a given data representation, some regions of the input space may be classified confidently, while other regions may be ambiguous.}
    \label{fig:motivation}
  \end{figure}

  With this in mind, we propose a mechanism for introducing cascaded evaluation to arbitrary feedforward ANNs, focusing on the task of object recognition as a proof of concept. Instead of classifying images only at the final layer, every layer in the network may attempt to classify images in low-ambiguity regions of its input space, passing ambiguous images forward to subsequent layers for further consideration (see Fig. \ref*{fig:motivation} for an illustration). We propose three approaches to training these networks, test them on small image datasets synthesized from MNIST \cite{lecun1998mnist} and CIFAR-10 \cite{krizhevsky2009learning}, and quantify the accuracy/efficiency trade-off that occurs when the network parameters are tuned to yield more aggressive early classification policies. Additionally, we propose and evaluate methods for appropriating regularization and optimization techniques developed for statically-routed networks.

  \section{Related Work}
  
  Since the late 1980s, researchers have combined artificial neural networks with decision trees in various ways \cite{utgoff1989perceptron} \cite{sirat1990neural}. More recently, \citet{kontschieder2015deep} performed joint optimization of ANN and decision tree parameters, and \citet{bulo2014neural} used randomized multi-layer networks to compute decision tree split functions.
    
  To our knowledge, the family of inference systems we discuss was first described by \citet{denoyer2014deep}. Additionally, \citet{bengio2015conditional} explored dynamically skipping layers in neural networks, and \citet{ioannou2016decision} explored dynamic routing in networks with equal-length paths. Some recently-developed visual detection systems perform cascaded evaluation of convolutional neural network layers \cite{li2015convolutional,cai2015learning,girshick2015fast,ren2015faster}; though highly specialized for the task of visual detection, these modifications can radically improve efficiency.
  
  While these approaches lend evidence that dynamic routing can be effective, they either ignore the cost of computation, or do not represent it explicitly, and instead use opaque heuristics to trade accuracy for efficiency. We build on this foundation by deriving training procedures from arbitrary application-provided costs of error and computation, comparing one actor-style and two critic-style strategies, and considering regularization and optimization in the context of dynamically-routed networks.

  \section{Setup}
  
  In a statically-routed, feedforward artificial neural network, every layer transforms a single input feature vector into a single output feature vector. The output feature vector is then used as the input to the following layer (which we'll refer to as the current layer's \textit{sink}), if it exists, or as the ouptut of the network as a whole, if it does not.
  
\begin{figure}[htb]
    \centering
    \includegraphics[width=\hsize]{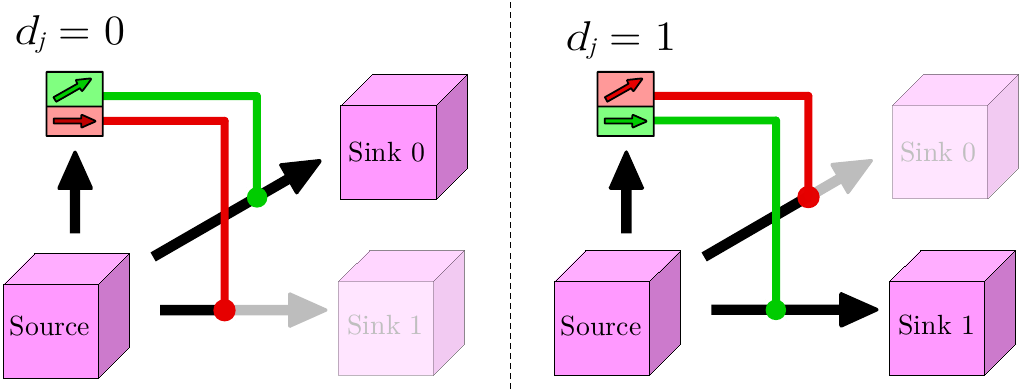}
    \caption{\textbf{A 2-way junction}, $j$. $d_j$ is an integer function of the source features. When $d_j=0$, the signal is propagated through the top sink, and the bottom sink is inactive. When $d_j=1$, the signal is propagated through the bottom sink, and the top sink is inactive.}
    \label{fig:junction}
  \end{figure}

  We consider networks in which layers may have more than one sink. In such a network, for every $n$-way junction $j$ a signal reaches, the network must make a decision, $d_j \in \{0..n\}$, such that the signal will propagate through the $i$\textsuperscript{th} sink if and only if $d_j = i$ (this is illustrated in Fig. \ref*{fig:junction}). We compute $d_j$ as the argmax of the score vector $s_j$, a learned function of the last feature vector computed before reaching $j$. We'll refer to this rule for generating $d$ from $s$ as the inference routing policy.
  
  \subsection{Multipath Architectures for Convolutional Networks}
  
  Convolutional network layers compute collections of \textit{local} descriptions of the input signal. It is unreasonable to expect that this kind of feature vector can explicitly encode the global information relevant to deciding how to route the entire signal (\textit{e.g.}, in the case of object recognition, whether the image was taken indoors, whether the image contains an animal, or the prevalence of occlusion in the scene).
  
\begin{figure*}[tb]
    \centering
    \includegraphics[width=\textwidth]{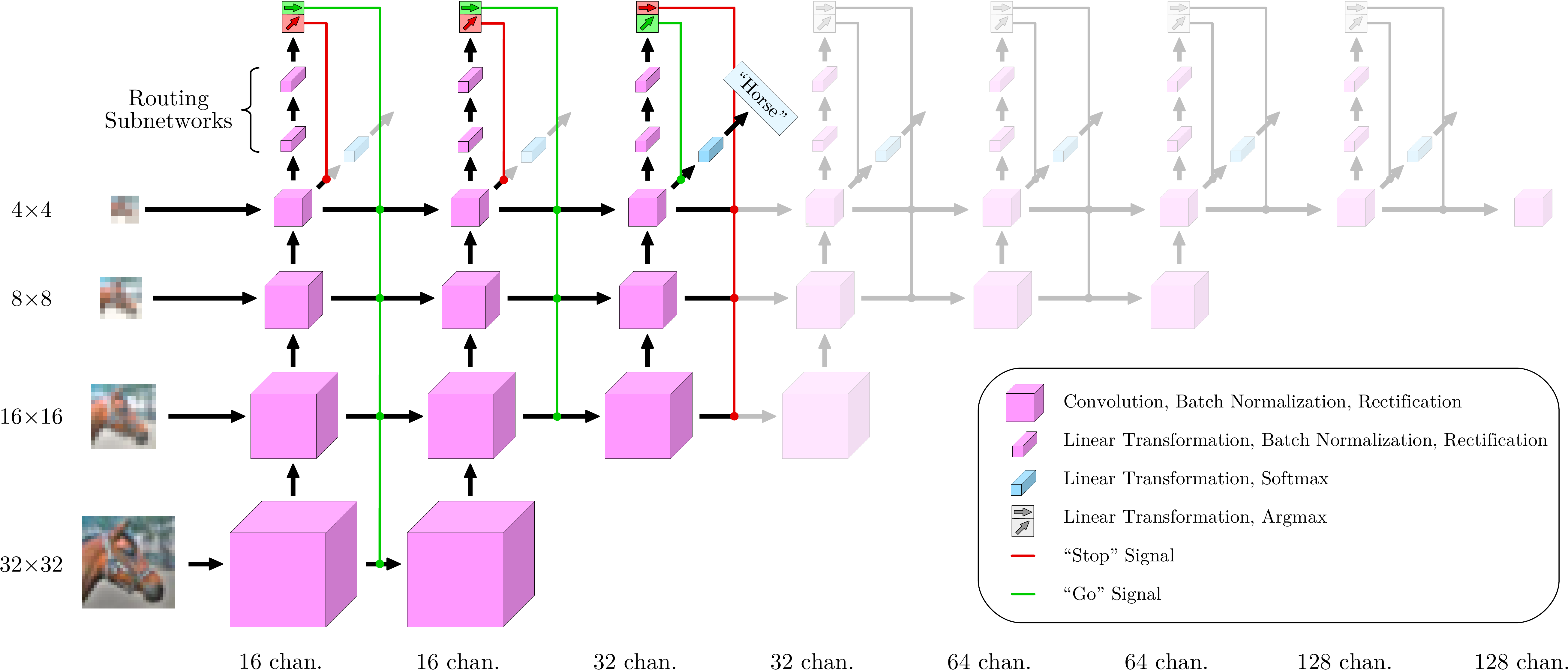}
    \caption{\textbf{Our multiscale convolutional architecture.} Once a column is evaluated, the network decides whether to classify the image or evaluate subsequent columns. Deeper columns operate at coarser scales, but compute higher-dimensional representations at each location. All convolutions use 3$\times$3 kernels, downsampling is achieved via 2$\times$2 max pooling, and all routing layers have 16 channels.}
    \label{fig:net}
  \end{figure*}

  To address this, instead of computing a 2-dimensional array of local features at each layer, we compute a pyramid of features (resembling the pyramids described by \citet{ke2016neural}), with local descriptors at the bottom and global descriptors at the top. At every junction $j$, the score vector $s_j$ is computed by a small routing network operating on the last-computed global descriptor. Our multipath architecture is illustrated in Fig. \ref*{fig:net}.

  \subsection{Balancing Accuracy and Efficiency}
  
  For a given input, network $\nu$, and set of routing decisions $d$, we define the cost of performing inference:
  
  \begin{equation}
    c_\text{inf}(\nu,d) = c_\text{err}(\nu,d) + c_\text{cpt}(\nu,d),
  \end{equation}
  
  where $c_\text{err}(\nu,d)$ is the cost of the inference errors made by the network, and $c_\text{cpt}(\nu,d)$ is the cost of computation. In our experiments, unless stated otherwise, $c_\text{err}$ is the cross-entropy loss and
  
  \begin{equation}
    c_\text{cpt}(\nu,d) = k_\text{cpt} n_\text{ops}(\nu,d),
  \end{equation}
  
  where $n_\text{ops}(\nu,d)$ is the number of multiply-accumulate operations performed and $k_\text{cpt}$ is a scalar hyperparameter. This definition assumes a time- or energy-constrained system---every operation consumes roughly the same amount of time and energy, so every operation is equally expensive. $c_\text{cpt}$ may be defined differently under other constraints (\textit{e.g.} memory bandwidth).

  \section{Training}

  We propose three approaches to training dynamically-routed networks, along with complementary approaches to regularization and optimization, and a method for adapting to changes in the cost of computation.

  \subsection{Training Strategy I: Actor Learning}
  \label{subesc:actor-learning}
  
  Since $d$ is discrete, $c_\text{inf}(\nu,d)$ cannot be minimized via gradient-based methods. However, if $d$ is replaced by a stochastic approximation, $\hat{d}$, during training, we can engineer the gradient of $\mathrm{E}[c_\text{inf}(\nu,\hat{d})]$ to be nonzero. We can then learn the routing parameters and classification parameters simultaneously by minimizing the loss
  
  \begin{equation}
    L_\text{ac} = \mathrm{E}[c_\text{inf}(\nu,\hat{d})].
  \end{equation}

  In our experiments, the training routing policy samples $\hat{d}$ such that
  
  \begin{equation}
    \Pr(\hat{d}_j=i) = \text{softmax}(s_j/\tau)_i,
  \end{equation}

  where $\tau$ is the network ``temperature'': a scalar hyperparameter that decays over the course of training, converging the training routing policy towards the inference routing policy.
  
  \subsection{Training Strategy II: Pragmatic Critic Learning}
  \label{subsec:pragmatic-critic-learning}

  Alternatively, we can attempt to learn to predict the expected utility of making every routing decision. In this case, we minimize the loss

  \begin{equation}
    L_\text{cr} = \mathrm{E} \left[
    c_\text{inf}(\nu,\hat{d}) + \sum_{j\in\mathcal{J}} c_\text{ure}^j \right],
  \end{equation}
  
  where $\mathcal{J}$ is the set of junctions encountered when making the routing decisions $\hat{d}$, and $c_\text{ure}$ is the utility regression error cost, defined:
  
  \begin{equation}
    c_\text{ure}^j = k_\text{ure} \|s_j - u_j\|^2,
  \end{equation}
  
  where
  
  \begin{equation}
    u_j^i = -c_\text{inf}(\nu_j^i,d),
  \end{equation}
  
  $k_\text{ure}$ is a scalar hyperparameter, and $\nu_j^i$ is the subnetwork consisting of the $i^\text{th}$ child of $\nu_j$, and all of its descendants. Since we want to learn the policy indirectly (via cost prediction), $\hat{d}$ is treated as constant with respect to optimization.

  \subsection{Training Strategy III: Optimistic Critic Learning}
  
  To improve the stability of the loss and potentially accelerate training, we can adjust the routing utility function $u$ such that, for every junction $j$, $u_j$ is independent of the routing parameters downstream of $j$. Instead of predicting the cost of making routing decisions given the \textit{current} downstream routing policy, we can predict the cost of making routing decisions given the \textit{optimal} downstream routing policy. In this optimistic variant of the critic method,

  \begin{equation}
    u_j^i = -\text{min}_{d'}(c_\text{inf}(\nu_j^i,d')).
  \end{equation}

  \subsection{Regularization}
  \label{subsec:regularization}

  Many regularization techniques involve adding a model-complexity term, $c_\text{mod}$, to the loss function to influence learning, effectively imposing soft constraints upon the network parameters \cite{hoerl1970ridge,rudin1992nonlinear,tibshirani1996regression}. However, if such a term affects layers in a way that is independent of the amount of signal routed through them, it will either underconstrain frequently-used layers or overconstrain infrequently-used layers. To support both frequently- and infrequently-used layers, we regularize subnetworks as they are activated by $\hat{d}$, instead of regularizing the entire network directly.
  
  For example, to apply L2 regularization to critic networks, we define $c_\text{mod}$:
  
  \begin{equation}
    c_\text{mod} = E \left[
      k_\text{L2} \sum_{w\in\mathcal{W}} w^2
    \right],
  \end{equation}
  
  where $\mathcal{W}$ is the set of weights associated with the layers activated by $\hat{d}$, and $k_\text{L2}$ is a scalar hyperparameter.
    
  For actor networks, we apply an extra term to control the magnitude of $s$, and therefore the extent to which the net explores subpotimal paths:
  
  \begin{equation}
    c_\text{mod} = E \left[
      k_\text{L2} \sum_{w\in\mathcal{W}} w^2 +
      k_\text{dec} \sum_{j\in\mathcal{J}} \|s_j\|^2
    \right],
  \end{equation}
  
  where $k_\text{dec}$ is a scalar hyperparameter indicating the relative cost of decisiveness.

  $c_\text{mod}$ is added to the loss function in all of our experiments. Within $c_\text{mod}$, unless stated otherwise, $\hat{d}$ is treated as constant with respect to optimization.

  \subsection{Adjusting Learning Rates to Compensate for Throughput Variations}
  \label{subsec:optimization}
  
  Both training techniques attempt to minimize the expected cost of performing inference with the network, over the training routing policy. With this setup, if we use a constant learning rate for every layer in the network, then layers through which the policy routes examples more frequently will receive larger parameter updates, since they contribute more to the expected cost.

  To allow every layer to learn as quickly as possible, we scale the learning rate of each layer $\ell$ dynamically, by a factor $\alpha_\ell$, such that the elementwise variance of the loss gradient with respect to $\ell$'s parameters is independent of the amount of probability density routed through it.
 
  To derive $\alpha_\ell$, we consider an alternative routing policy, $d_\ell^\ast$, that routes all signals though $\ell$, then routes through subsequent layers based on $\hat{d}$. With this policy, at every training interation, mini-batch stochastic gradient descent shifts the parameters associated with layer $\ell$ by a vector $\delta_\ell^\ast$, defined:

  \begin{equation}
    \delta_\ell^\ast = -\lambda \sum_i g_\ell^i,
  \end{equation}
  
  where $\lambda$ is the global learning rate and $g_\ell^i$ is the gradient of the loss with respect to the parameters in $\ell$, for training example $i$, under $d_\ell^\ast$. Analogously, the scaled parameter adjustment under $\hat{d}$ can be written

  \begin{equation}
    \delta_\ell = -\alpha_\ell \lambda \sum_i p_\ell^i g_\ell^i,
  \end{equation}
  
  where $p_\ell^i$ is the probability with which $\hat{d}$ routes example $i$ through $\ell$.
  
  We want to select $\alpha_\ell$ such that
  
  \begin{equation}
    \mathrm{Var}(\delta_\ell) =
    \mathrm{Var}(\delta_\ell^\ast).
  \end{equation}

  Substituting the definitions of $\delta_\ell$ and $\delta_\ell^\ast$,

  \begin{equation}
    \mathrm{Var} \left( \alpha_\ell \sum_i p_\ell^i g_\ell^i \right) =
    \mathrm{Var} \left( \sum_i g_\ell^i \right).
  \end{equation}
  
  Since every $g_\ell^i$ is sampled independently, we can rewrite this equation:
  
  \begin{equation}
    n_\text{ex} v_\ell \alpha_\ell^2 \|p_\ell\|^2 =
    n_\text{ex} v_\ell,
  \end{equation}

  where $n_\text{ex}$ is the number of training examples in the mini-batch and $v_\ell$ is the elementwise variance of $g_\ell^i$, for any $i$ (since every example is sampled via the same mechanism). We can now show that

  \begin{equation}
    \alpha_\ell = \|p_\ell\|^{-1}.
  \end{equation}

  So, for every layer $\ell$, we can scale the learning rate by $\|p_\ell\|^{-1}$, and the variance of the weight updates will be similar thoughout the network. We use this technique, unless otherwise specified, in all of our experiments.
  
  \subsection{Responding to Changes in the Cost of Computation}
  \label{subsec:dyn-kcpt}

  We may want a single network to perform well in situations with various degrees of computational resource scarcity (\textit{e.g.} computation may be more expensive when a device battery is low). To make the network's routing behavior responsive to a dynamic $c_\text{cpt}$, we can concatenate $c_\text{cpt}$'s known parameters---in our case, $\{k_\text{cpt}\}$---to the input of every routing subnetwork, to allow them to modulate the routing policy. To match the scale of the image features and facilitate optimization, we express $k_\text{cpt}$ in units of cost per ten-million operations.

  \subsection{Hyperparameters}
  
  In all of our experiments, we use a mini-batch size, $n_\text{ex}$, of 128, and run 80,000 training iterations. We perform stochastic gradient descent with initial learning rate $0.1/n_\text{ex}$ and momentum 0.9. The learning rate decays continuously with a half-life of 10,000 iterations.
  
  The weights of the final layers of routing networks are zero-initialized, and we initialize all other weights using the Xavier initialization method \cite{glorot2010understanding}. All biases are zero-initialized. We perform batch normalization \cite{ioffe2015batch} before every rectification operation, with an $\epsilon$ of $1\times10^{-6}$, and an exponential moving average decay constant of 0.9.
  
  $\tau$ is initialized to 1.0 for actor networks and 0.1 for critic networks, and decays with a half-life of 10,000 iterations. $k_\text{dec} = 0.01$, $k_\text{ure} = 0.001$, and $k_\text{L2} = 1\times10^{-4}$. We selected these values (for $\tau$, $k_\text{dec}$, $k_\text{ure}$, and $k_\text{L2}$) by exploring the hyperparameter space logarithmically, by powers of 10, training and evaluating on the hybrid MNIST/CIFAR-10 dataset (described in section \ref*{subsec:policy-learning}). At a coarse level, these values are locally optimal---multiplying or dividing any of them by 10 will not improve performance.

  \subsection{Data Augmentation}

  We augment our data using an approach that is popular for use with CIFAR-10 \cite{lin2013network} \cite{srivastava2015training} \cite{clevert2015fast}. We augment each image by applying vertical and horizontal shifts sampled uniformly from the range [-4px,4px], and, if the image is from CIFAR-10, flipping it horizontally with probability 0.5. We fill blank pixels introduced by shifts with the mean color of the image (after gamma-decoding).
  
  \section{Experiments}
  \label{sec:experiments}

\begin{figure*}[p]
    \centering
    \includegraphics[width=\textwidth]{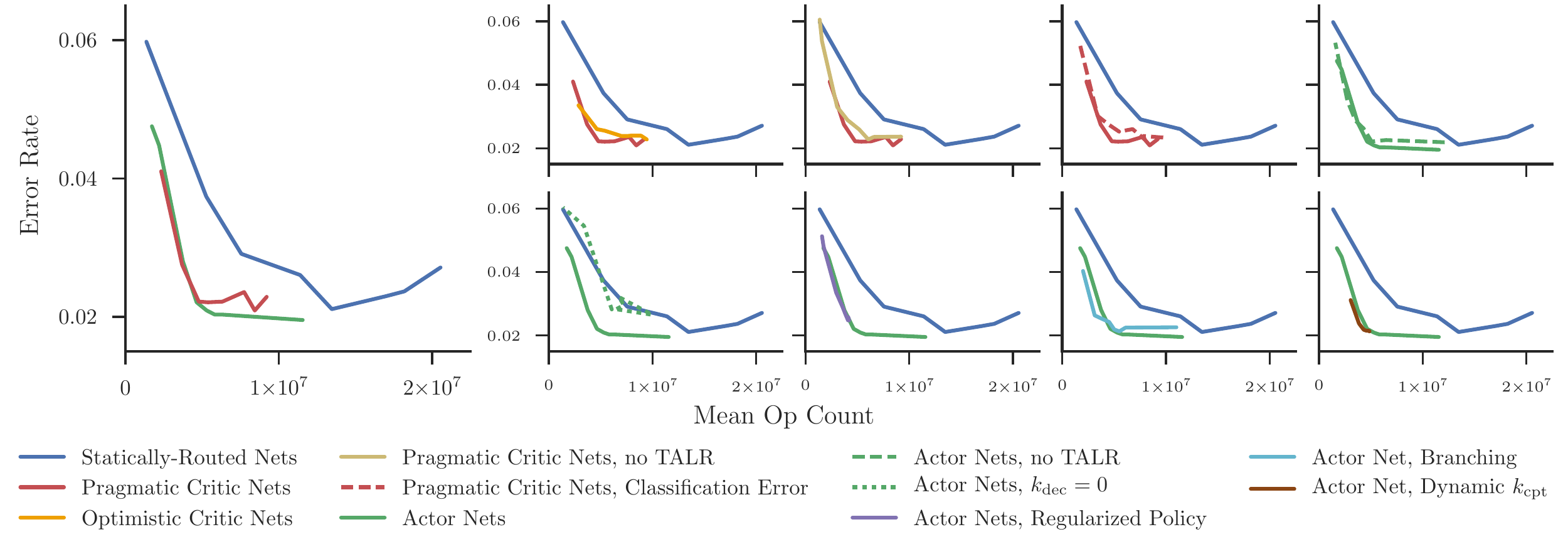}
    \caption{\textbf{Hybrid dataset performance.} Every point along the ``statically-routed nets'' curve corresponds to a network composed of the first $n$ columns of the architecture illustrated in Fig. \ref*{fig:net}, for $1 \leq n \leq 8$. The points along the ``actor net, dynamic $k_\text{cpt}$'' curve correspond to a single network evaluated with various values of $k_\text{cpt}$, as described in section \ref*{subsec:dyn-kcpt}. The points along all other curves correspond to distinct networks, trained with different values of $k_\text{cpt}$. $k_\text{cpt} \in \{0, \num{1e-9}, \num{2e-9}, \num{4e-9}, ...~ \num{6.4e-8}\}$.}
    \label{fig:acc-eff-00}
  \end{figure*}

\begin{figure*}[p]
    \centering
    \includegraphics[width=\hsize]{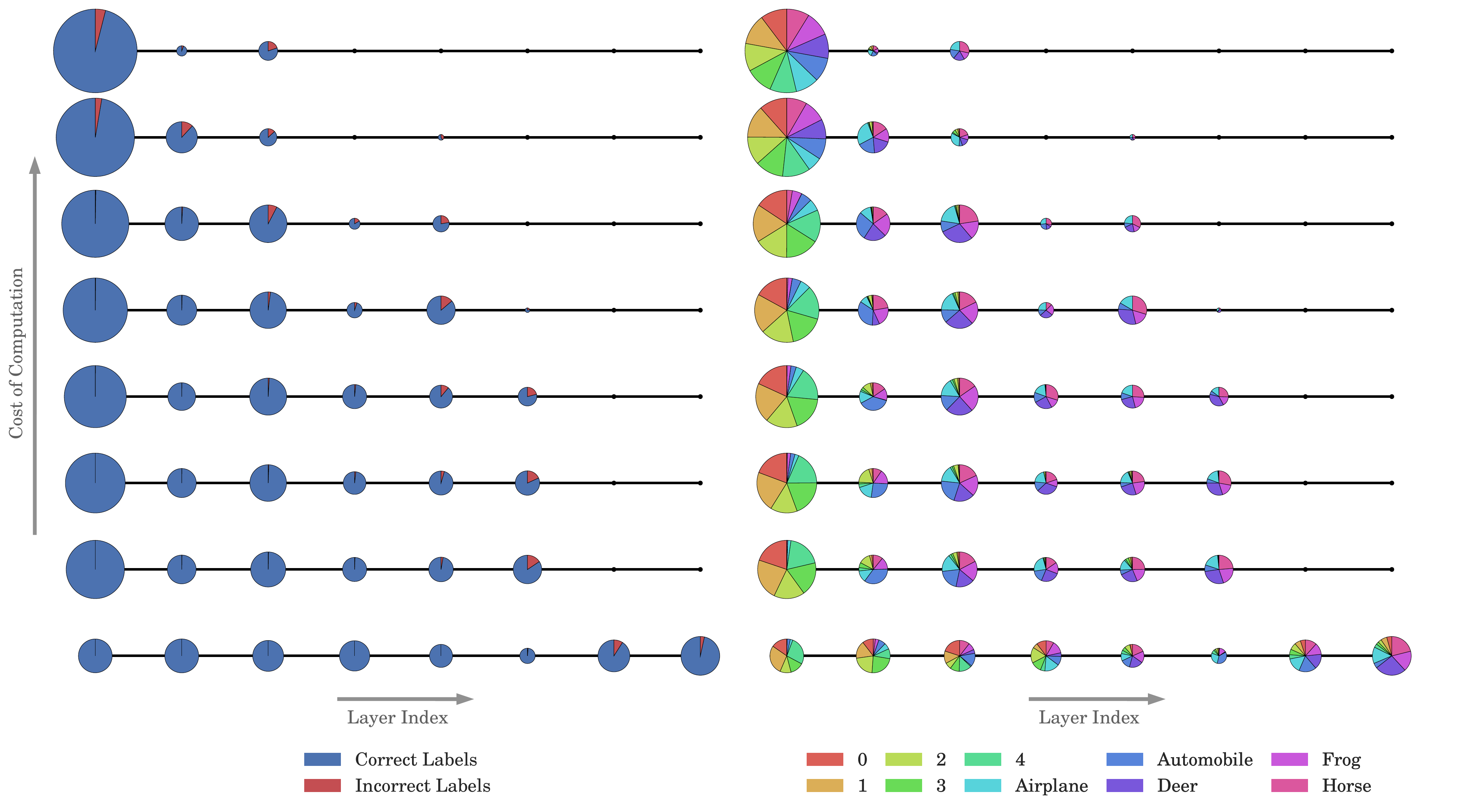}
    \caption{\textbf{Dataflow  through  actor networks} trained to classify images from the hybrid MNIST/CIFAR-10 dataset. Every row is a node-link diagram corresponding to a network, trained with a different $k_\text{cpt}$. Each circle indicates,  by area,  the fraction of examples that are classified at the corresponding layer. The circles are colored to indicate the accuracy of each layer (left) and the kinds of images classified at each layer (right).}
    \label{fig:nld-00}
  \end{figure*}

\begin{figure}[htb]
    \centering
    \includegraphics[width=\hsize]{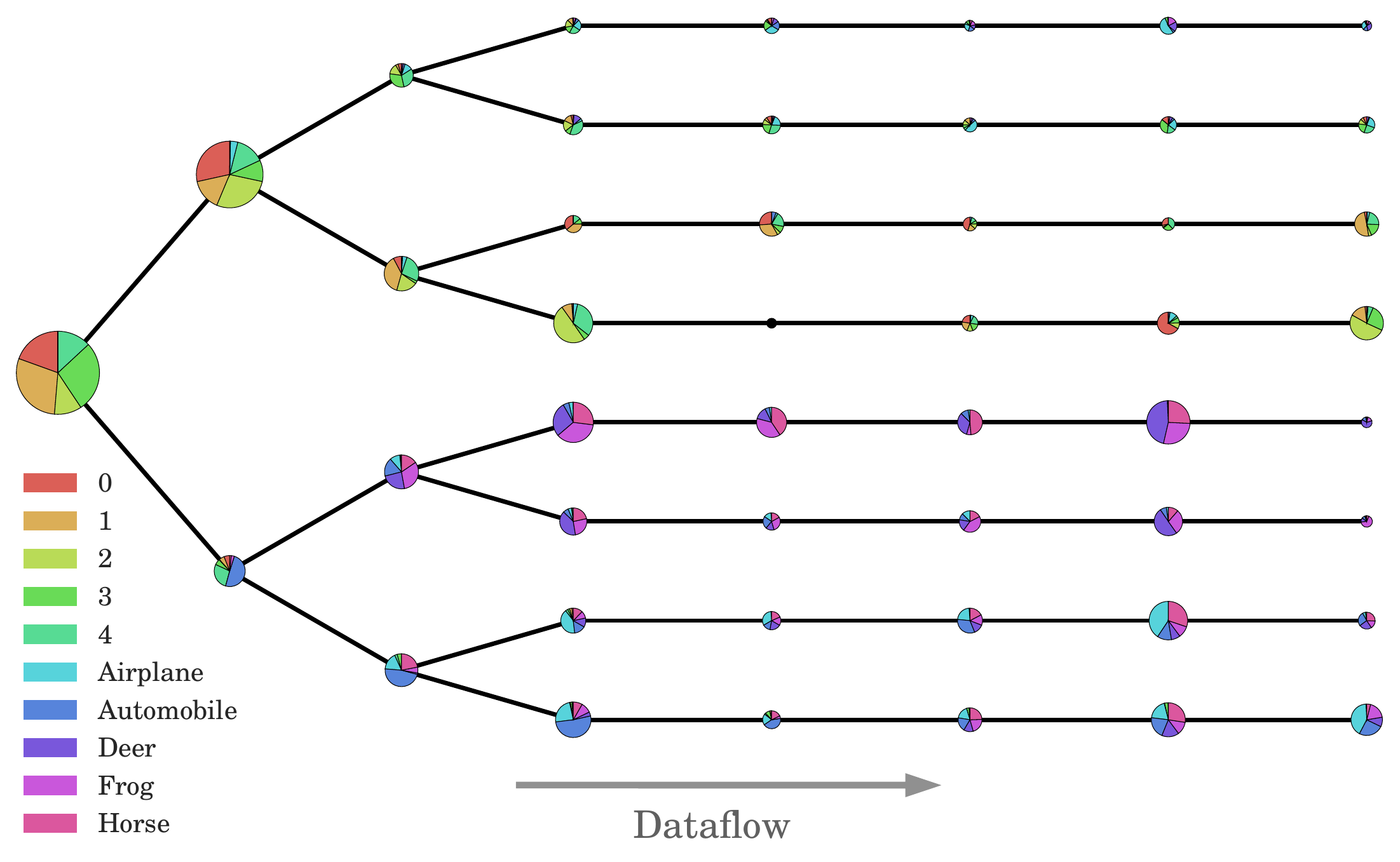}
    \caption{\textbf{Dataflow  through  a branching actor network} trained to classify images in the hybrid dataset, illustrated as in Fig. \ref*{fig:nld-00}.}
    \label{fig:nld-01}
  \end{figure}

  We compare approaches to dynamic routing by training 153 networks to classify small images, varying the policy-learning strategy, regularization strategy, optimization strategy, architecture, cost of computation, and details of the task. The results of these experiments are reported in Fig. \ref*{fig:acc-eff-00}--\ref*{fig:acc-eff-02}. Our code is available \href{https://github.com/MasonMcGill/multipath-nn}{on GitHub}.

  \subsection{Comparing Policy-Learning Strategies}
  \label{subsec:policy-learning}
  
\begin{figure}[htb]
    \centering
    \includegraphics[width=\hsize]{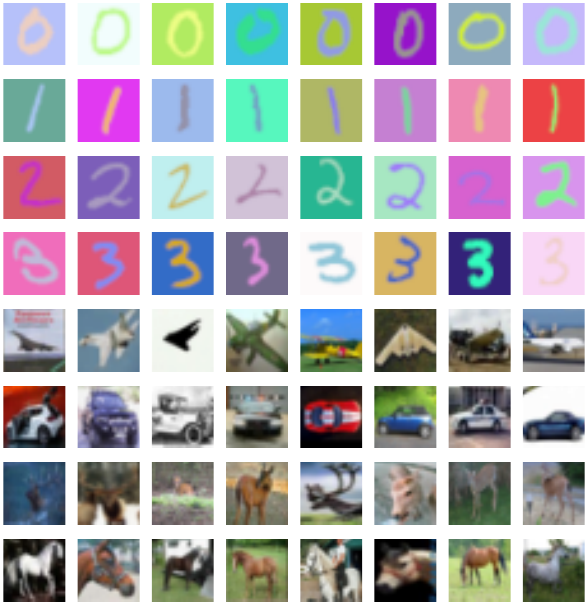}
    \caption{\textbf{Sample images from the hybrid MNIST/CIFAR-10 dataset.} We recolor images from MNIST via the following procedure: we select two random colors at least 0.3 units away from each other in RGB space; we then map black pixels to the first color, map white pixels to the second color, and linearly interpolate in between.}
    \label{fig:dataset}
  \end{figure}

  To compare routing strategies in the context of a simple dataset with a high degree of difficulty variation, we train networks to classify images from a small-image dataset synthesized from MNIST \cite{lecun1998mnist} and CIFAR-10 \cite{krizhevsky2009learning} (see Fig. \ref*{fig:dataset}). Our dataset includes the classes ``0'', ``1'', ``2'', ``3'', and ``4'' from MNIST and ``airplane'', ``automobile'', ``deer'', ``horse'', and ``frog'' from CIFAR-10 (see Fig. \ref*{fig:dataset}). The images from MNIST are resized to match the scale of images from CIFAR-10 (32$\times$32), via linear interpolation, and are color-modulated to make them more difficult to trivially distinguish from CIFAR-10 images (MNIST is a grayscale dataset).
  
\begin{figure}[htb]
    \centering
    \includegraphics[width=\hsize]{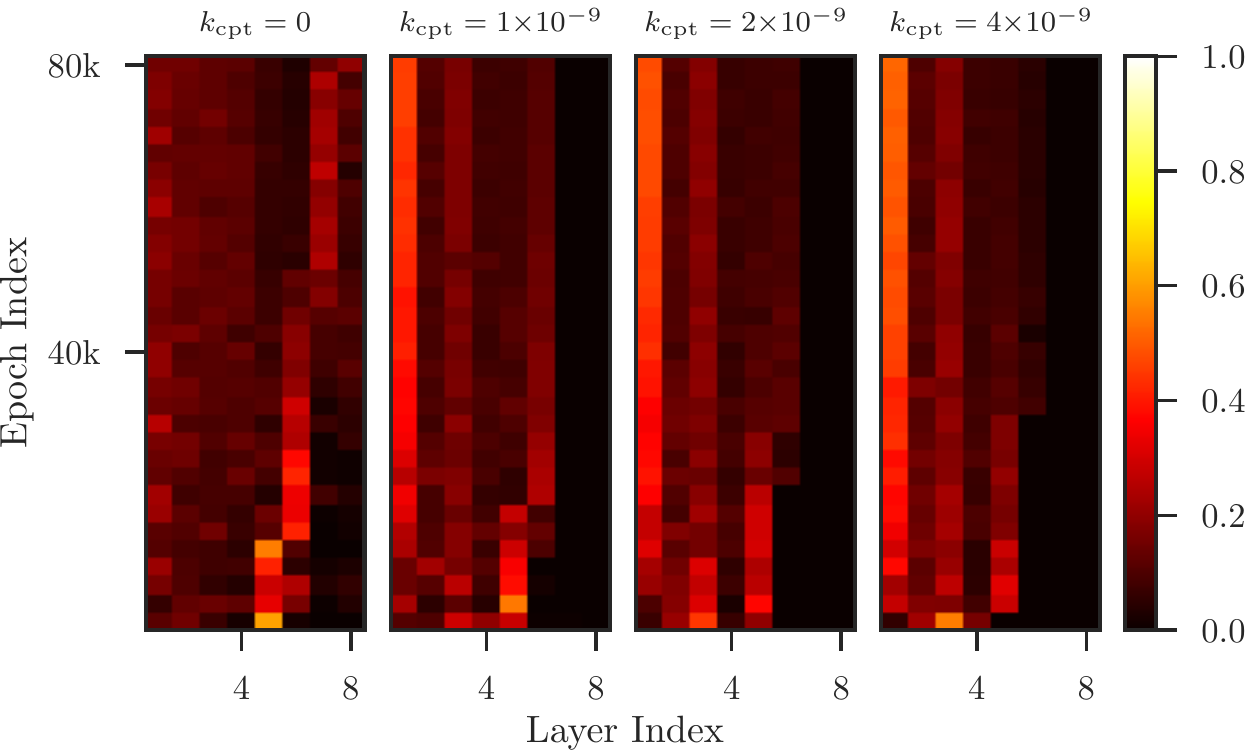}
    \caption{\textbf{Dataflow over the course of training.} The heatmaps illustrate the fraction of validation images classified at every terminal node in the bottom four networks in Fig. \ref*{fig:nld-00}, over the course of training.}
    \label{fig:routing-hists}
  \end{figure}

  For a given computational budget, dynamically-routed networks achieve higher accuracy rates than architecture-matched statically-routed baselines (networks composed of the first $n$ columns of the architecture illustrated in Fig. \ref*{fig:net}, for $n \in \{1..8\}$). Additionally, dynamically-routed networks tend to avoid routing data along deep paths at the beginning of training (see Fig. \ref*{fig:routing-hists}). This is possibly because the error surfaces of deeper networks are more complicated, or because deeper paths are less stable---changing the parameters in any component layer to better classify images routed along other, overlapping paths may decrease performance. Whatever the mechanism, this tendency to initially find simpler solutions seems to prevent some of the overfitting that occurs with 7- and 8-layer statically-routed networks.
  
  Compared to other dynamically-routed networks, optimistic critic networks perform poorly, possibly because optimal routers are a poor approximation for our small, low-capacity router networks. Actor networks perform better than critic networks, possibly because critic networks are forced to learn a potentially-intractable auxilliary task (\text{i.e.} it's easier to decide who to call to fix your printer than it is to predict exactly how quickly and effectively everyone you know would fix it). Actor networks also consistently achieve higher peak accuracy rates than comparable statically-routed networks, across experiments.
  
  Although actor networks may be more performant, critic networks are more flexible. Since critic networks don't require $\mathrm{E}[c_\text{inf}(\nu,\hat{d})]$ to be a differentiable function of $\hat{d}$, they can be trained by sampling $\hat{d}$, saving memory, and they support a wider selection of training routing policies (\text{e.g.} $\epsilon$-greedy) and $c_\text{inf}$ definitions. In addition to training the standard critic networks, we train networks using a variant of the pragmatic critic training policy, in which we replace the cross-entropy error in the $c_\text{ure}$ term with the classification error. Although these networks do not perform as well as the original pragmatic critic networks, they still outperform comparable statically-routed networks.

  \subsection{Comparing Regularization Strategies}

  Based on our experiments with the hybrid dataset, regularizing $\hat{d}$, as described in section \ref*{subsec:regularization}, discourages networks from routing data along deep paths, reducing peak accuracy. Additionally, some mechanism for encouraging exploration (in our case, a nonzero $k_\text{dec}$) appears to be necessary to train effective actor networks.
  
  \subsection{Comparing Optimization Strategies}

  Throughput-adjusting the learning rates (TALR), as described in section \ref*{subsec:optimization}, improves the hybrid dataset performance of both actor and critic networks in computational-resource-abundant, high-accuracy contexts.
  
  \subsection{Comparing Architectures}

  For a given computational budget, architectures with both 2- and 3-way junctions have a higher capacity than subtrees with only 2-way junctions. On the hybrid dataset, under tight computational constraints, we find that trees with higher degrees of branching achieve higher accuracy rates. Unconstrained, however, they are prone to overfitting.
  
  In dynamically-routed networks, early classification layers tend to have high accuracy rates, pushing difficult decisions downstream. Even without energy contraints, terminal layers specialize in detecting instances of certain classes of images. These classes are usually related (they either all come from MNIST or all come from CIFAR-10.) In networks with both 2- and 3-way junctions, branches specialize to an even greater extent. (See Fig. \ref*{fig:nld-00} and \ref*{fig:nld-01}.)
  
  \subsection{Comparing Specialized and Adaptive Networks}
  \label{subsec:multi-purpose-nets}

  We train a single actor network to classify images from the hybrid datset under various levels of computational constraints, using the approach described in section \ref*{subsec:dyn-kcpt}, sampling $k_\text{cpt}$ randomly from the set mentioned in Fig. \ref*{fig:acc-eff-00} for each training example. This network performs comparably to a collection of 8 actor nets trained with various static values of $k_\text{cpt}$, over a significant, central region of the accuracy/efficiency curve, with an 8-fold reduction in memory consumption and training time.

  \subsection{Exploring the Effects of the Decision Difficulty Distribution}
  \label{subsec:diff-dist}
  
\begin{figure}[htb]
    \centering
    \includegraphics[width=\hsize]{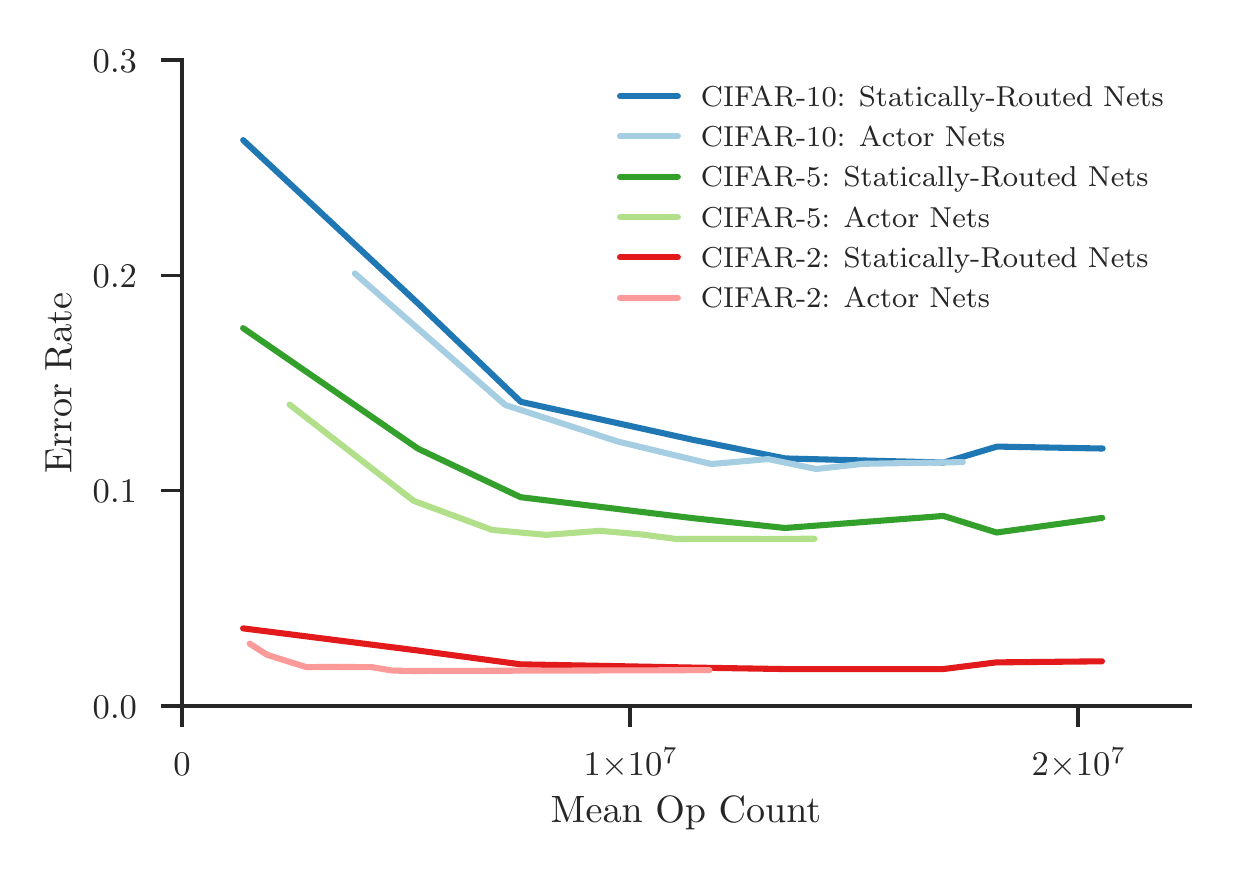}
    \caption{\textbf{Performance effects of the task difficulty distribution}, as described in section \ref*{subsec:diff-dist}. The ``statically-routed nets'' and ``actor nets'' curves are drawn analogously to their counterparts in Fig. \ref*{fig:acc-eff-00}.}
    \label{fig:acc-eff-01}
  \end{figure}

  To probe the effect of the inference task's difficulty distribution on the performance of dynamically-routed networks, we train networks to classify images from CIFAR-10, adjusting the classification task to vary the frequency of difficult decisions (see Fig. \ref*{fig:acc-eff-01}). We call these variants CIFAR-2---labelling images as ``horse'' or ``other''---and CIFAR-5---labelling images  as ``cat'', ``dog'', ``deer'', ``horse'', or ``other''. In this experiment, we compare actor networks (the best-performing networks from the first set of experiments) to architecture-matched statically-routed networks.
 
  \begin{figure}[htb]
    \centering
    \includegraphics[width=\hsize]{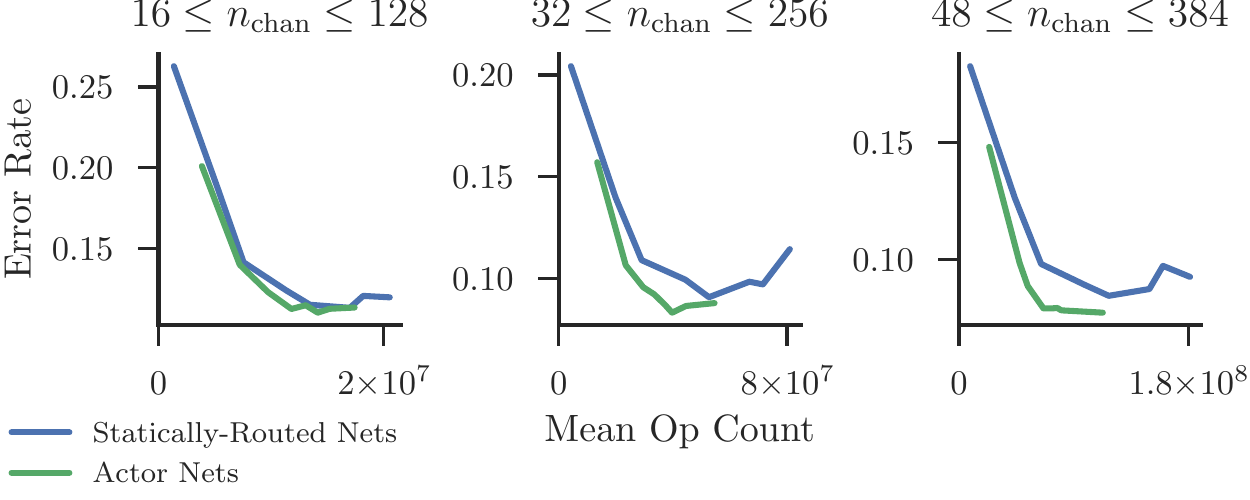}
    \caption{\textbf{Performance effects of model capacity}, training and testing on CIFAR-10. (Left) Networks with (subsets of) the architecture illustrated in Fig. \ref*{fig:net}. (Center) Networks otherwise identical to those presented in the left panel, with the number of output channels of every convolutional layer multiplied by 2, and $k_\text{cpt}$ divided by 4. (Right) Networks otherwise identical to those presented in the left panel, with the number of output channels of every convolutional layer multiplied by 3, and $k_\text{cpt}$ divided by 9.}
    \label{fig:acc-eff-02}
  \end{figure}
 
  We find that dynamic routing is more beneficial when the task involves many low-difficulty decisions, allowing the network to route more data along shorter paths. While dynamic routing offers only a slight advantage on CIFAR-10, dynamically-routed networks achieve a higher peak accuracy rate on CIFAR-2 than statically-routed networks, at a third of the computational cost.

  \subsection{Exploring the Effects of Model Capacity}
  
  To test whether dynamic routing is advantageous in higher-capacity settings, we train actor networks and architecture-matched statically-routed networks to classify images from CIFAR-10, varying the width of the networks (see Fig. \ref*{fig:acc-eff-02}). Increasing the model capacity either increases or does not affect the relative advantage of dynamically-routed networks, suggesting that our approach is applicable to more complicated tasks.

  \section{Discussion}
  
  Our experiments suggest that dynamically-routed networks trained under mild computational constraints can operate 2--3 times more efficiently than comparable statically-routed networks, without sacrificing performance. Additionally, despite their higher capacity, dynamically-routed networks may be less prone to overfitting.
  
  When designing a multipath architecture, we suggest supporting early decision-making wherever possible, since cheap, simple routing networks seem to work well. In convolutional architectures, pyramidal layers appear to be reasonable sites for branching.
  
  The actor strategy described in section \ref*{subesc:actor-learning} is generally an effective way to learn a routing policy. However, the pragmatic critic strategy described in section \ref*{subsec:pragmatic-critic-learning} may be better suited for very large networks (trained via decision sampling to conserve memory) or networks designed for applications with nonsmooth cost-of-inference functions---\textit{e.g.} one in which $k_\text{cpt}$ has units \textit{errors$/$operation}. Adjusting learning rates to compensate for throughput variations, as described in section \ref*{subsec:optimization}, may improve the performance of deep networks. If the cost of computation is dynamic, a single network, trained with the procedure described in section \ref*{subsec:multi-purpose-nets}, may still be sufficient.
  
  While we test our approach on tasks with some degree of difficulty variation, it is possible that dynamic routing is even more advantageous when performing more complex tasks. For example, video annotation may require specialized modules to recognize locations, objects, faces, human actions, and other scene components or attributes, but having every module constantly operating may be extremely inefficient. A dynamic routing policy could fuse these modules, allowing them to share common components, and activate specialized components as necessary.
  
  Another interesting topic for future research is growing and shrinking dynamically-routed networks during training. With such a network, it is not necessary to specify an architecture. The network will instead take shape over the course of training, as computational contraints, memory contraints, and the data dictate.

  \section*{Acknowledgements}
  
  This work was funded by a generous grant from Google Inc. We would also like to thank Krzysztof Chalupka, Cristina Segalin, and Oisin Mac Aodha for their thoughtful comments.

  \bibliography{references}

\begin{thebibliography}{32}
\providecommand{\natexlab}[1]{#1}
\providecommand{\url}[1]{\texttt{#1}}
\expandafter\ifx\csname urlstyle\endcsname\relax
  \providecommand{\doi}[1]{doi: #1}\else
  \providecommand{\doi}{doi: \begingroup \urlstyle{rm}\Url}\fi

\bibitem[Bengio et~al.(2015)Bengio, Bacon, Pineau, and
  Precup]{bengio2015conditional}
Bengio, Emmanuel, Bacon, Pierre-Luc, Pineau, Joelle, and Precup, Doina.
\newblock Conditional computation in neural networks for faster models.
\newblock \emph{arXiv preprint arXiv:1511.06297}, 2015.

\bibitem[Bulo \& Kontschieder(2014)Bulo and Kontschieder]{bulo2014neural}
Bulo, Samuel and Kontschieder, Peter.
\newblock Neural decision forests for semantic image labelling.
\newblock In \emph{Proceedings of the IEEE Conference on Computer Vision and
  Pattern Recognition}, pp.\  81--88, 2014.

\bibitem[Cai et~al.(2015)Cai, Saberian, and Vasconcelos]{cai2015learning}
Cai, Zhaowei, Saberian, Mohammad, and Vasconcelos, Nuno.
\newblock Learning complexity-aware cascades for deep pedestrian detection.
\newblock In \emph{Proceedings of the IEEE International Conference on Computer
  Vision}, pp.\  3361--3369, 2015.

\bibitem[Clevert et~al.(2015)Clevert, Unterthiner, and
  Hochreiter]{clevert2015fast}
Clevert, Djork-Arn{\'e}, Unterthiner, Thomas, and Hochreiter, Sepp.
\newblock Fast and accurate deep network learning by exponential linear units
  (elus).
\newblock \emph{arXiv preprint arXiv:1511.07289}, 2015.

\bibitem[Denoyer \& Gallinari(2014)Denoyer and Gallinari]{denoyer2014deep}
Denoyer, Ludovic and Gallinari, Patrick.
\newblock Deep sequential neural network.
\newblock \emph{arXiv preprint arXiv:1410.0510}, 2014.

\bibitem[Dosovitskiy et~al.(2015)Dosovitskiy, Fischer, Ilg, Hausser, Hazirbas,
  Golkov, van~der Smagt, Cremers, and Brox]{dosovitskiy2015flownet}
Dosovitskiy, Alexey, Fischer, Philipp, Ilg, Eddy, Hausser, Philip, Hazirbas,
  Caner, Golkov, Vladimir, van~der Smagt, Patrick, Cremers, Daniel, and Brox,
  Thomas.
\newblock Flownet: Learning optical flow with convolutional networks.
\newblock In \emph{Proceedings of the IEEE International Conference on Computer
  Vision}, pp.\  2758--2766, 2015.

\bibitem[Girshick(2015)]{girshick2015fast}
Girshick, Ross.
\newblock Fast r-cnn.
\newblock In \emph{Proceedings of the IEEE International Conference on Computer
  Vision}, pp.\  1440--1448, 2015.

\bibitem[Glorot \& Bengio(2010)Glorot and Bengio]{glorot2010understanding}
Glorot, Xavier and Bengio, Yoshua.
\newblock Understanding the difficulty of training deep feedforward neural
  networks.
\newblock In \emph{Aistats}, volume~9, pp.\  249--256, 2010.

\bibitem[Goodale \& Milner(1992)Goodale and Milner]{goodale1992separate}
Goodale, Melvyn~A and Milner, A~David.
\newblock Separate visual pathways for perception and action.
\newblock \emph{Trends in neurosciences}, 15\penalty0 (1):\penalty0 20--25,
  1992.

\bibitem[He et~al.(2016)He, Zhang, Ren, and Sun]{he2016deep}
He, Kaiming, Zhang, Xiangyu, Ren, Shaoqing, and Sun, Jian.
\newblock Deep residual learning for image recognition.
\newblock In \emph{Proceedings of the IEEE Conference on Computer Vision and
  Pattern Recognition}, pp.\  770--778, 2016.

\bibitem[Ho(1995)]{ho1995random}
Ho, Tin~Kam.
\newblock Random decision forests.
\newblock In \emph{Document Analysis and Recognition, 1995., Proceedings of the
  Third International Conference on}, volume~1, pp.\  278--282. IEEE, 1995.

\bibitem[Hoerl \& Kennard(1970)Hoerl and Kennard]{hoerl1970ridge}
Hoerl, Arthur~E and Kennard, Robert~W.
\newblock Ridge regression: Biased estimation for nonorthogonal problems.
\newblock \emph{Technometrics}, 12\penalty0 (1):\penalty0 55--67, 1970.

\bibitem[Ioannou et~al.(2016)Ioannou, Robertson, Zikic, Kontschieder, Shotton,
  Brown, and Criminisi]{ioannou2016decision}
Ioannou, Yani, Robertson, Duncan, Zikic, Darko, Kontschieder, Peter, Shotton,
  Jamie, Brown, Matthew, and Criminisi, Antonio.
\newblock Decision forests, convolutional networks and the models in-between.
\newblock \emph{arXiv preprint arXiv:1603.01250}, 2016.

\bibitem[Ioffe \& Szegedy(2015)Ioffe and Szegedy]{ioffe2015batch}
Ioffe, Sergey and Szegedy, Christian.
\newblock Batch normalization: Accelerating deep network training by reducing
  internal covariate shift.
\newblock \emph{arXiv preprint arXiv:1502.03167}, 2015.

\bibitem[Ke et~al.(2016)Ke, Maire, and Yu]{ke2016neural}
Ke, Tsung-Wei, Maire, Michael, and Yu, Stella~X.
\newblock Neural multigrid.
\newblock \emph{arXiv preprint arXiv:1611.07661}, 2016.

\bibitem[Kontschieder et~al.(2015)Kontschieder, Fiterau, Criminisi, and
  Rota~Bulo]{kontschieder2015deep}
Kontschieder, Peter, Fiterau, Madalina, Criminisi, Antonio, and Rota~Bulo,
  Samuel.
\newblock Deep neural decision forests.
\newblock In \emph{Proceedings of the IEEE International Conference on Computer
  Vision}, pp.\  1467--1475, 2015.

\bibitem[Kornblith et~al.(2013)Kornblith, Cheng, Ohayon, and
  Tsao]{kornblith2013network}
Kornblith, Simon, Cheng, Xueqi, Ohayon, Shay, and Tsao, Doris~Y.
\newblock A network for scene processing in the macaque temporal lobe.
\newblock \emph{Neuron}, 79\penalty0 (4):\penalty0 766--781, 2013.

\bibitem[Krizhevsky \& Hinton(2009)Krizhevsky and
  Hinton]{krizhevsky2009learning}
Krizhevsky, Alex and Hinton, Geoffrey.
\newblock Learning multiple layers of features from tiny images.
\newblock 2009.

\bibitem[LeCun et~al.(1998)LeCun, Cortes, and Burges]{lecun1998mnist}
LeCun, Yann, Cortes, Corinna, and Burges, Christopher~JC.
\newblock The mnist database of handwritten digits, 1998.

\bibitem[Li et~al.(2015)Li, Lin, Shen, Brandt, and Hua]{li2015convolutional}
Li, Haoxiang, Lin, Zhe, Shen, Xiaohui, Brandt, Jonathan, and Hua, Gang.
\newblock A convolutional neural network cascade for face detection.
\newblock In \emph{Proceedings of the IEEE Conference on Computer Vision and
  Pattern Recognition}, pp.\  5325--5334, 2015.

\bibitem[Lin et~al.(2013)Lin, Chen, and Yan]{lin2013network}
Lin, Min, Chen, Qiang, and Yan, Shuicheng.
\newblock Network in network.
\newblock \emph{arXiv preprint arXiv:1312.4400}, 2013.

\bibitem[Moeller et~al.(2008)Moeller, Freiwald, and Tsao]{moeller2008patches}
Moeller, Sebastian, Freiwald, Winrich~A, and Tsao, Doris~Y.
\newblock Patches with links: a unified system for processing faces in the
  macaque temporal lobe.
\newblock \emph{Science}, 320\penalty0 (5881):\penalty0 1355--1359, 2008.

\bibitem[Newell et~al.(2016)Newell, Yang, and Deng]{newell2016stacked}
Newell, Alejandro, Yang, Kaiyu, and Deng, Jia.
\newblock Stacked hourglass networks for human pose estimation.
\newblock In \emph{European Conference on Computer Vision}, pp.\  483--499.
  Springer, 2016.

\bibitem[Ren et~al.(2015)Ren, He, Girshick, and Sun]{ren2015faster}
Ren, Shaoqing, He, Kaiming, Girshick, Ross, and Sun, Jian.
\newblock Faster r-cnn: Towards real-time object detection with region proposal
  networks.
\newblock In \emph{Advances in Neural Information Processing Systems}, pp.\
  91--99, 2015.

\bibitem[Rudin et~al.(1992)Rudin, Osher, and Fatemi]{rudin1992nonlinear}
Rudin, Leonid~I, Osher, Stanley, and Fatemi, Emad.
\newblock Nonlinear total variation based noise removal algorithms.
\newblock \emph{Physica D: Nonlinear Phenomena}, 60\penalty0 (1-4):\penalty0
  259--268, 1992.

\bibitem[Simonyan \& Zisserman(2014)Simonyan and Zisserman]{simonyan2014very}
Simonyan, Karen and Zisserman, Andrew.
\newblock Very deep convolutional networks for large-scale image recognition.
\newblock \emph{arXiv preprint arXiv:1409.1556}, 2014.

\bibitem[Sirat \& Nadal(1990)Sirat and Nadal]{sirat1990neural}
Sirat, JA and Nadal, JP.
\newblock Neural trees: a new tool for classification.
\newblock \emph{Network: Computation in Neural Systems}, 1\penalty0
  (4):\penalty0 423--438, 1990.

\bibitem[Srivastava et~al.(2015)Srivastava, Greff, and
  Schmidhuber]{srivastava2015training}
Srivastava, Rupesh~K, Greff, Klaus, and Schmidhuber, J{\"u}rgen.
\newblock Training very deep networks.
\newblock In \emph{Advances in neural information processing systems}, pp.\
  2377--2385, 2015.

\bibitem[Tibshirani(1996)]{tibshirani1996regression}
Tibshirani, Robert.
\newblock Regression shrinkage and selection via the lasso.
\newblock \emph{Journal of the Royal Statistical Society. Series B
  (Methodological)}, pp.\  267--288, 1996.

\bibitem[Utgoff(1989)]{utgoff1989perceptron}
Utgoff, Paul~E.
\newblock Perceptron trees: A case study in hybrid concept representations.
\newblock \emph{Connection Science}, 1\penalty0 (4):\penalty0 377--391, 1989.

\bibitem[Viola et~al.(2005)Viola, Jones, and Snow]{viola2005detecting}
Viola, Paul, Jones, Michael~J, and Snow, Daniel.
\newblock Detecting pedestrians using patterns of motion and appearance.
\newblock \emph{International Journal of Computer Vision}, 63\penalty0
  (2):\penalty0 153--161, 2005.

\bibitem[Zhou et~al.(2013)Zhou, Fan, Cao, Jiang, and Yin]{zhou2013extensive}
Zhou, Erjin, Fan, Haoqiang, Cao, Zhimin, Jiang, Yuning, and Yin, Qi.
\newblock Extensive facial landmark localization with coarse-to-fine
  convolutional network cascade.
\newblock In \emph{Proceedings of the IEEE International Conference on Computer
  Vision Workshops}, pp.\  386--391, 2013.

\end{thebibliography}
  \bibliographystyle{icml2017}
\end{document}